\documentclass[lettersize,journal]{IEEEtran}
\usepackage{float}
\usepackage{amsmath,amsfonts}
\usepackage{algorithmic}
\usepackage{algorithm}
\usepackage{array}
\usepackage[caption=false,font=normalsize,labelfont=sf,textfont=sf]{subfig}
\usepackage{textcomp}
\usepackage{stfloats}
\usepackage{url}
\usepackage{xurl}
\usepackage{verbatim}
\usepackage{subcaption}
\usepackage{graphicx}
\usepackage{caption}
\usepackage{xcolor}
\usepackage{makecell}
\usepackage{multirow}
\usepackage{tablefootnote}
\usepackage{multicol}
\usepackage{latexsym}
\usepackage{booktabs}
\usepackage{tabularx} 
\usepackage{cite}
\usepackage[
    colorlinks=true,
    citecolor=blue,
    linkcolor=black,
    urlcolor=blue
]{hyperref}
\usepackage{orcidlink}
\newcolumntype{Y}{>{\raggedright\arraybackslash}X}

\begin{document}

\title{A Scaled Three-Vehicle Platooning Platform}

\author{Kaiyue Lu\textsuperscript{1}, Qiaoxuan Zhang\textsuperscript{1}, Yukun Lu\textsuperscript{1*}\,\orcidlink{0000-0003-3259-5945}

\textsuperscript{1} Dept. of Mechanical Engineering, University of New Brunswick, Fredericton, NB E3B 5A3, Canada

}

\maketitle

\section{Introduction}
Vehicle platooning has attracted increasing attention as a promising approach to improve traffic efficiency, energy consumption, and roadway safety through coordinated multi-vehicle operation. A key challenge in platooning lies in maintaining stable and accurate path tracking during dynamic maneuvers such as lane changes, where lateral deviations and heading disturbances generated by the lead vehicle may propagate downstream to following vehicles. Robust longitudinal and lateral control systems are therefore essential not only for individual vehicle tracking performance, but also for overall platoon stability.

Among practical path-tracking methods, the PID, Pure Pursuit, and Stanley controllers are widely adopted because of their intuitive structure, low computational demand, and suitability for real-time implementation~\cite{Snider2009, park-pp, sunstanley}. 
In this study, the Pure Pursuit and Stanley methods are implemented as lateral tracking controllers and experimentally evaluated during a lane-change maneuver, respectively. A classic PID controller is utilized for longitudinal control~\cite{pid}. Performance is assessed in terms of following distance, lateral tracking accuracy, velocity regulation, yaw response, and disturbance propagation across the platoon.

For experimental studies, the Intelligent Mobility and Robotics Lab (IMRL) develops a scaled multi-vehicle platform for autonomous platooning research, with a particular emphasis on cooperative control and human-in-the-loop autonomy~\cite{lu2026rationale}. This platform consists of one human-operable lead vehicle and two autonomous followers, enabling controlled and repeatable experiments on leader-follower coordination. Compared with full-scale field testing, this scaled platform offers a safer, lower-cost, and more flexible environment for rapid prototyping, controller validation, and multi-agent autonomy studies, while providing stronger physical realism than purely simulation-based evaluations~\cite{shu2026simulation}.

\begin{figure}[ht]
    \centering
    \includegraphics[width=0.7\linewidth]{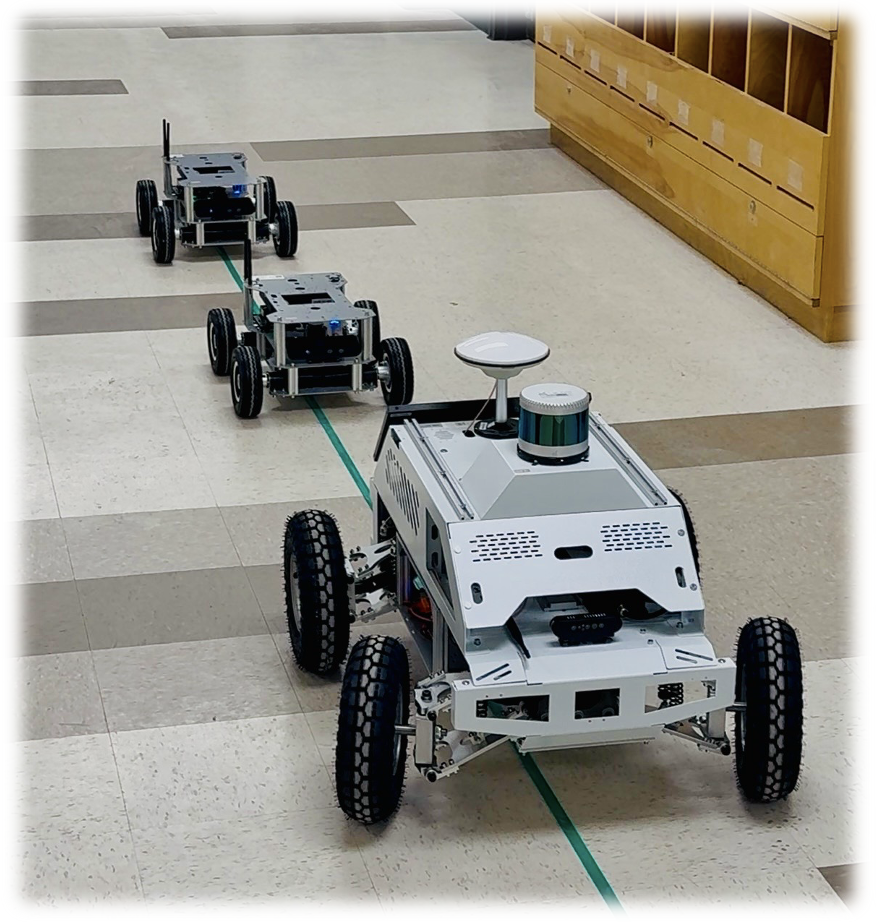}
    \caption{The scaled vehicle platform at the Intelligent Mobility and Robotics Lab (IMRL).}
\end{figure}

\section{Scaled Multi-Vehicle Platform}
\label{platform}
The platooning platform comprises one lead vehicle and two follower vehicles. The lead vehicle employs a front-wheel Ackermann steering configuration, while each follower vehicle adopts a four-wheel differential-drive architecture. All vehicles are equipped with an NVIDIA Jetson Orin NX 16~GB module for high-level onboard computation and an STM32F407VET6 microcontroller for low-level control and actuator interfacing.

The Jetson Orin module operates on Ubuntu 22.04 with ROS 2 Humble~\cite{ros2_humble_docs}. It is responsible for perception and high-level decision-making tasks, which demand substantial computational resources but comparatively modest real-time requirements. The Jetson Orin processes multi-sensor data streams, including camera and LiDAR measurements, together with IMU and wheel-odometry data received from the STM32 controller. Based on the fused sensor information, it generates high-level control commands, such as the desired longitudinal velocity and steering angle, which are subsequently transmitted to the STM32 board for low-level actuation.

The STM32 board operates under FreeRTOS~\cite{freertos} and is responsible for executing the control commands generated by the Jetson Orin. In addition, it acquires IMU measurements and wheel-odometry data, which are subsequently transmitted to the Jetson Orin through a UART-TTL communication interface. Because the STM32 outputs UART logic-level signals, a UART-TTL conversion module is employed to ensure voltage compatibility and to protect the Jetson Orin from improper signal levels.


Note that the actuators and sensor configurations differ between the lead vehicle and the follower vehicles, they are described separately in the following sections.

\begin{table}[ht]
    \centering
    \caption{Nomenclature}
    \begin{tabular}{ll}
        \toprule
        \textbf{Symbol} & \textbf{Description} \\
        \midrule
        $L$ & Wheelbase (m) \\
        $W$ & Track-width (m) \\
        $\psi$ & Yaw angle (rad) \\
        $\delta$ & Steering angle (rad) \\
        
        $\delta_l$ & Front-left wheel steering angle (rad) \\
        $\delta_r$ & Front-right wheel steering angle (rad) \\
        $v_{x,f}$ & Longitudinal velocity of the follower vehicle (m/s) \\
        
        $v_{x,\mathrm{obj}}$ & Objective longitudinal velocity (m/s) \\
        $v_{y,\mathrm{obj}}$ & Objective lateral velocity (m/s) \\
        $\delta_\mathrm{obj}$ & Objective steering angle (rad) \\
        $\delta_{l,\mathrm{obj}}$ & Objective front-left wheel steering angle (rad) \\
        
        $\hat\delta_l$ & Measured front-left steering angle from motor encoder (rad) \\

        $\hat{a}_x$ & Measured longitudinal acceleration from 6-axis IMU (m/s\textsuperscript{2}) \\
        $\hat{a}_y$ & Measured lateral acceleration from 6-axis IMU (m/s\textsuperscript{2}) \\
        $\hat{a}_z$  & Measured vertical acceleration from 6-axis IMU (m/s\textsuperscript{2}) \\
        $\hat \omega$ & Measured yaw rate from 6-axis IMU (rad/s) \\
        $\hat{\dot\varphi}$ & Measured roll rate from 6-axis IMU (rad/s) \\
        $\hat{\dot\theta}$ & Measured pitch rate from 6-axis IMU (rad/s) \\
        
        $d_{\text{measure}}$ & Measured inter-vehicle distance from perception system (m) \\
        $d_{\text{goal}}$ & Desired inter-vehicle distance (m) \\

        $\hat v_{x}$ & Estimated longitudinal velocity (m/s) by Eqs.~\ref{eq:vx_leader} and~\ref{eq:vx_follower} \\
        $\hat v_y$ & Estimated lateral velocity (m/s) by Eq.~\ref{eq:vy_leader} \\
        
        $\alpha$ & Line-of-Sight (LOS) angle (rad) \\
        $l_d$ & Lookahead distance (m) \\
        $K_{l_d}$ & Lookahead gain \\
        $K_p$ & Proportional gain \\
        $K_i$ & Integral gain \\
        $K_d$ & Derivative gain \\
        $K_y$ & Crosstrack gain \\
        $e_y$ & Crosstrack error (m) \\
        $e_x$ & Inter-vehicle distance tracking error (m) \\
        $e_\psi$ & Inter-vehicle heading error (rad) \\
        $\varepsilon_v$ & Regularization constant \\ 
        \bottomrule
    \end{tabular}
    \label{tab:nomenclature}
\end{table}

\subsection{Lead Vehicle}
The lead vehicle is a four-wheel mobile platform configured with Ackermann steering. Steering is realized through a servo motor that controls the orientation of the two front wheels, while propulsion is provided by the two rear drive wheels. Each rear wheel is actuated by a 24~V DC motor equipped with a 500 pulses-per-second GMR encoder. The motors have a maximum rotational speed of 175~rpm and a peak torque of 0.557~kgm. An electric motor driver regulates the voltage supplied to the two rear motors based on PWM signals generated by the STM32 board. Encoder feedback from the rear motors is simultaneously acquired and used to estimate wheel speed and vehicle odometry.

\begin{figure}[ht]
    \centering
    \includegraphics[width=\linewidth]{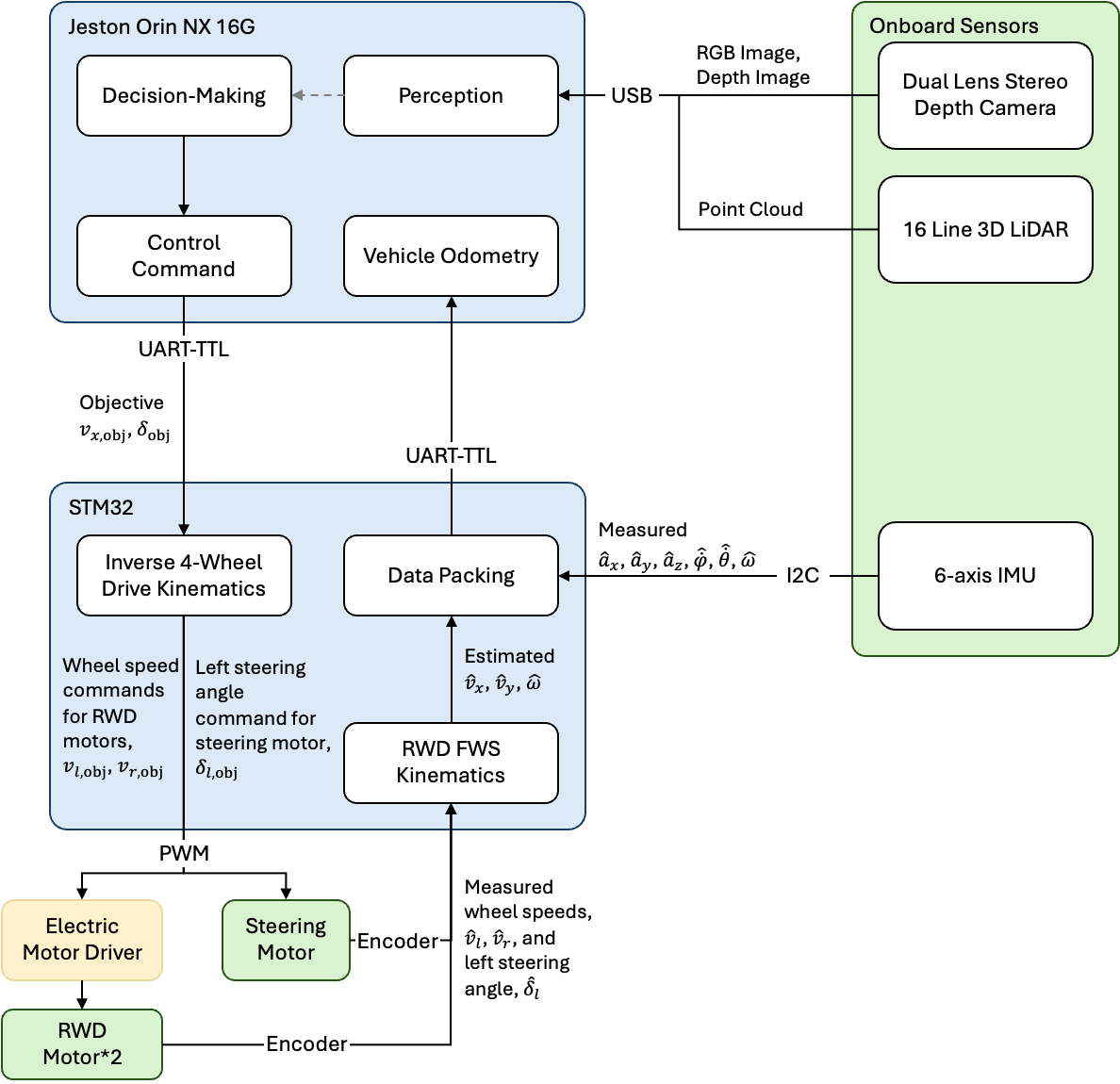}
    \caption{Lead vehicle hardware architecture.}
    \label{fig:hardware-ackerman}
\end{figure}

The front-facing camera mounted on the lead vehicle is an ORBBEC Gemini Pro, a V4L2-compatible stereo depth camera. As a Video4Linux2 (V4L2) device, it is natively recognized by the Linux operating system through the V4L2 framework, enabling direct access to image streams by software packages such as OpenCV, ROS, and standard camera-viewer applications. The camera is connected to the Jetson Orin via a USB-C interface and employs a dual-lens stereo vision architecture.
The depth stream provides a resolution of 1280 × 800 at 30 fps and is transmitted through the OpenNI 2.0 API. The effective sensing range is 0.2–2.5 m, with an accuracy of approximately 5 mm at a distance of 1 m under indoor operating conditions. The horizontal and vertical fields of view for depth sensing are 67.9° and 45.3°, respectively.
In addition, the RGB image stream provides a resolution of 1280 × 720 at 30 fps and is transmitted through the driver-free UVC (USB Video Class) protocol. The horizontal and vertical fields of view of the RGB camera are 71.5° and 56.7°, respectively.

The LiDAR sensor is an LSLIDAR C16, a 16-channel three-dimensional LiDAR unit. It supports scanning frequencies of up to 20 Hz and generates point-cloud data at rates of up to 640,000 points per second. The vertical field of view ranges from -16° to +14°. Point-cloud data are transmitted to the Jetson Orin board via an Ethernet interface.

\begin{figure}[ht]
    \centering
    \includegraphics[width=\linewidth]{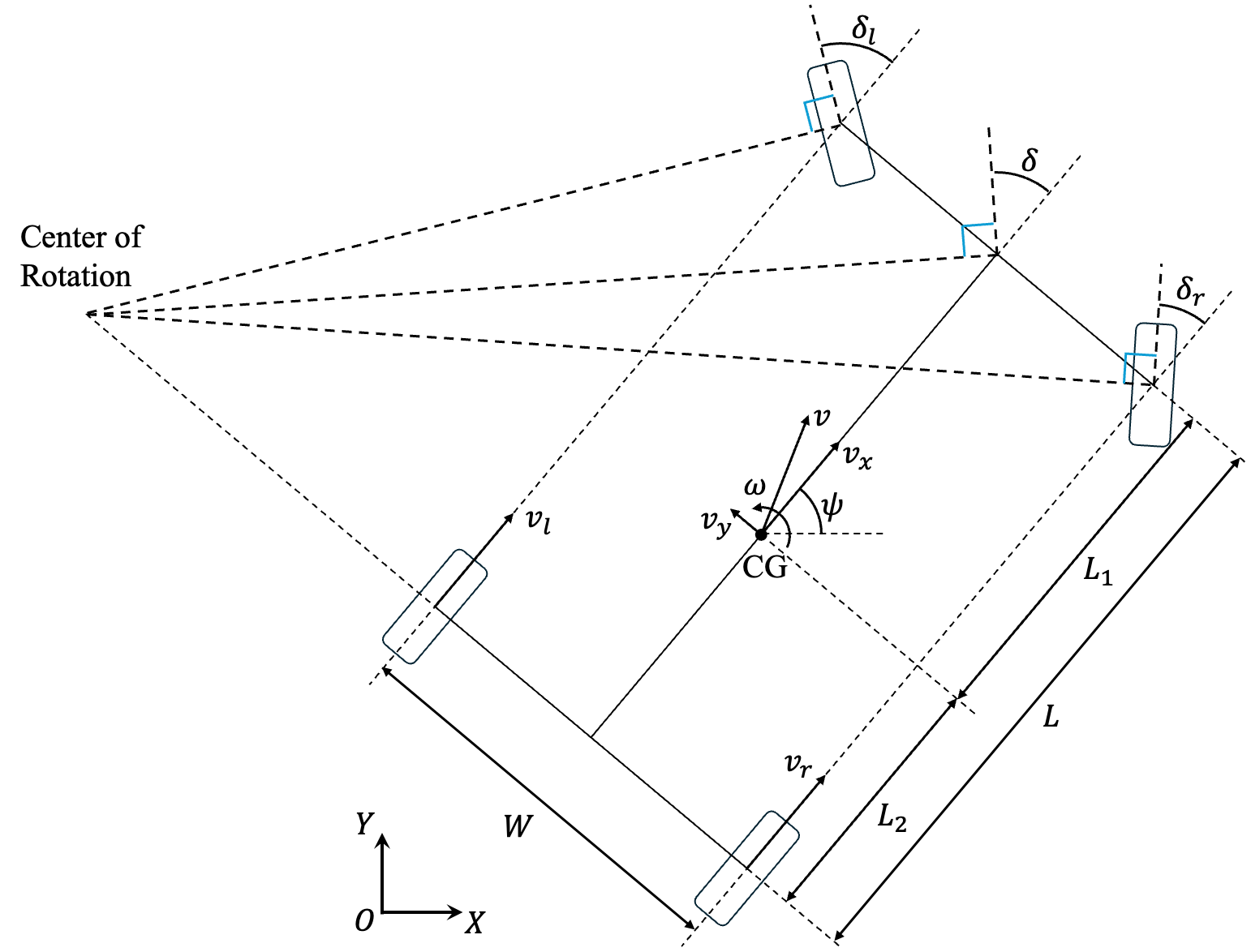}
    \caption{Schematic of the lead vehicle with Ackermann steering and rear-wheel drive systems.}
    \label{fig:vehicle-ackerman}
\end{figure}

The IMU used in the lead vehicle is the ICM20948, which is integrated into the STM32 board. The IMU incorporates a three-axis accelerometer and a three-axis gyroscope. The sensor transmits three-axis acceleration data in m/s\textsuperscript{2} and three-axis angular velocity data in rad/s to the STM32 board through the I2C communication interface.

The STM32 functions as the low-level controller, actuating the objective vehicle motions by controlling individual wheel speeds. 
The full chassis control of the lead vehicle is achieved by setting the objective rear-left wheel speed $v_{l,\mathrm{obj}}$, rear-right wheel speed $v_{r,\mathrm{obj}}$, and front-left wheel steering angle $\delta_{l,\mathrm{obj}}$.
These control variables can be calculated using the Inverse Ackermann kinematics model~\cite{ackermankinematics} by giving the objective longitudinal velocity $v_{x,\mathrm{obj}}$ and vehicle steering angle $\delta_\mathrm{obj}$:


\begin{equation}
    v_{l,\mathrm{obj}} = v_{x,\mathrm{obj}} \left(1 - \frac{W}{2L}\tan\delta_\mathrm{obj}\right)
\end{equation}
\begin{equation}
    v_{r,\mathrm{obj}} = v_{x,\mathrm{obj}} \left(1 + \frac{W}{2L}\tan\delta_\mathrm{obj}\right)
\end{equation}
\begin{equation}
    \delta_{l,\mathrm{obj}} = \mathrm{atan}\frac{2L\tan\delta_\mathrm{obj}}{2L - W\tan\delta_\mathrm{obj}}
\end{equation}
where $v_{x,\mathrm{obj}}$ is the objective longitudinal velocity, $W$ is the track-width, and $L$ is the wheelbase.

As shown in Figure~\ref{fig:vehicle-ackerman}, the front-right wheel steering angle $\delta_r$ can be then determined under the Ackermann condition via
\begin{equation}
    \cot\delta_r - \cot\delta_{l,\mathrm{obj}} = \frac{W}{L}
\end{equation}

Based on the Ackermann kinematic model, the vehicle’s longitudinal velocity $\hat v_x$, lateral velocity $\hat v_y$, and yaw rate $\hat \omega$ can be estimated from wheel-speed measurements obtained from the motor encoders together with the front-axle steering angle provided by the steering actuator, as follows:
\begin{equation}
    \hat v_x = \frac{\hat v_l + \hat v_r}{2}
    \label{eq:vx_leader}
\end{equation}
\begin{equation}
    \hat v_y = \frac{L_2}{L\cot\hat\delta_l + \frac{W}{2}} \cdot \hat v_x
    \label{eq:vy_leader}
\end{equation}
\begin{equation}
    \hat \omega = \frac{\hat v_r - \hat v_l}{W}
\end{equation}
where $\hat v_l$ is the measured rear-left wheel speed, $\hat v_r$ is the measured rear-right wheel speed, $\hat \delta_l$ is the measured front-left wheel steering angle, and $L_2$ is the distance between the rear axle and the vehicle CG.

\subsection{Follower Vehicle}
The follower vehicle is a four-wheel-drive (4WD) mobile platform with a differential-steering configuration. Steering is realized by regulating the speed difference between the left and right wheels, while propulsion is provided by all four wheels. Each wheel is actuated by a 24~V brushless DC motor equipped with a 500 pulses-per-second GMR encoder. The motors have a maximum rotational speed of 310 rpm and a peak torque of 0.010~kgm. An electric motor driver regulates the voltage supplied to the four motors based on PWM signals generated by the STM32 board. Encoder feedback from all four motors is simultaneously acquired and used to estimate wheel speed and vehicle odometry.

\begin{figure}[ht]
    \centering
    \includegraphics[width=\linewidth]{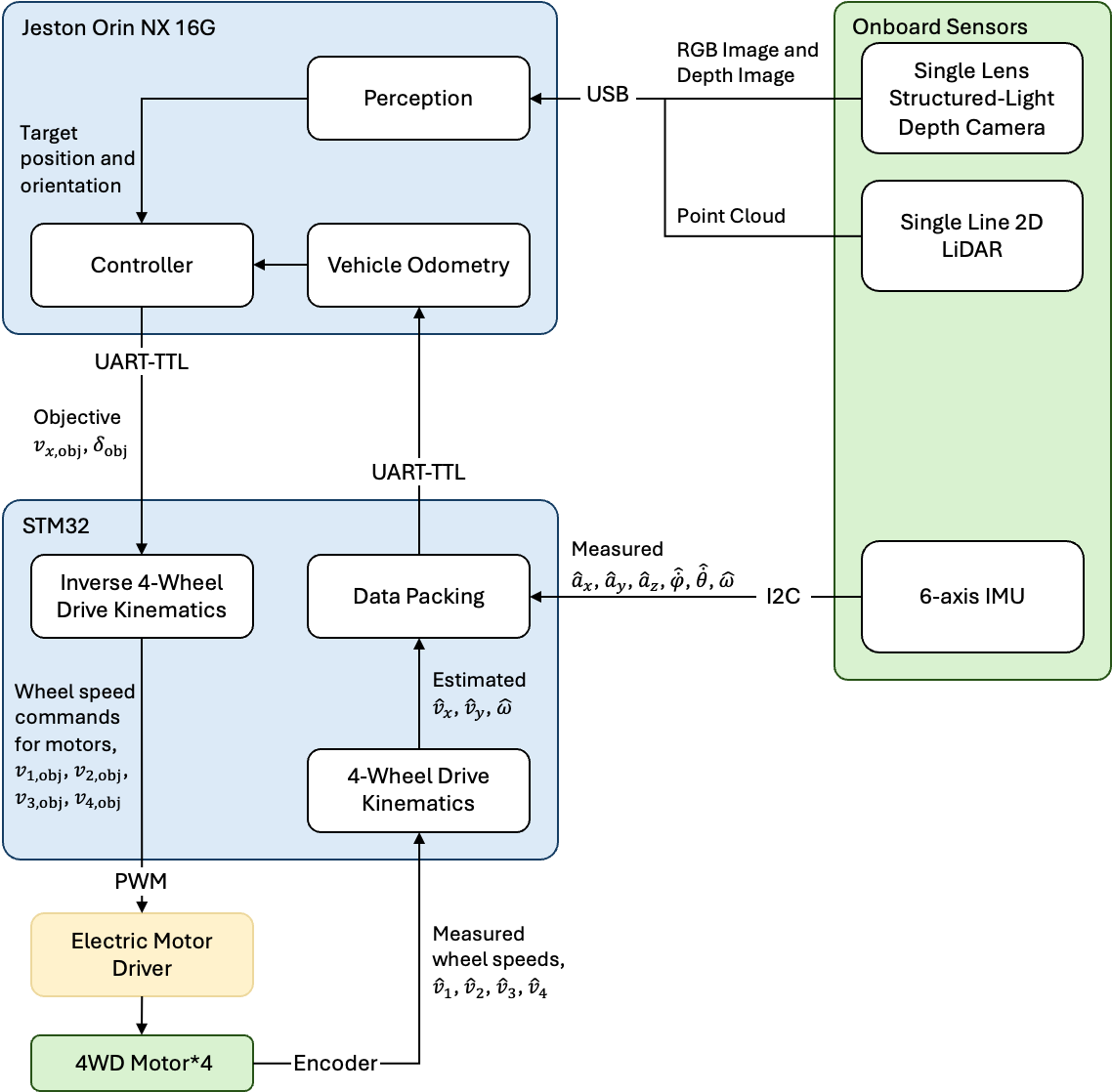}
    \caption{Follower vehicle hardware architecture.}
    \label{fig:hardware-4wd}
\end{figure}

The front-facing camera mounted on the follower vehicle is an ORBBEC Astra S, which is a standard V4L2-compatible device. It is a single-lens structured-light depth camera that employs an 850~nm infrared projector and is connected to the Jetson Orin through a USB-C interface.
The depth stream provides a resolution of 640 × 480 at 30 fps and is transmitted through the OpenNI 2.0 API. The effective sensing range is specified for indoor environments. The horizontal and vertical fields of view for depth sensing are 58.4° and 45.5°, respectively.
The RGB image stream also provides a resolution of 640 × 480 at 30 fps and is transmitted using the driver-free UVC protocol. The horizontal and vertical fields of view of the RGB camera are 63.1° and 49.4°, respectively.

The LiDAR sensor mounted on the follower vehicle is an LSLIDAR N10P, a single-line two-dimensional LiDAR unit. It supports scanning frequencies ranging from 6 to 12 Hz and provides a sampling rate of 5,400 points per second. LiDAR scan data are transmitted to the Jetson Orin through a USB-A interface.

The IMU module used in the follower vehicle is also an ICM20948, identical to that equipped on the lead vehicle. The STM32 serves as the low-level controller, executing the commanded vehicle motions based on the platform kinematic model, which defines the relationship between individual wheel motions and overall vehicle movement.

\begin{figure}[ht]
    \centering
    \includegraphics[width=0.8\linewidth]{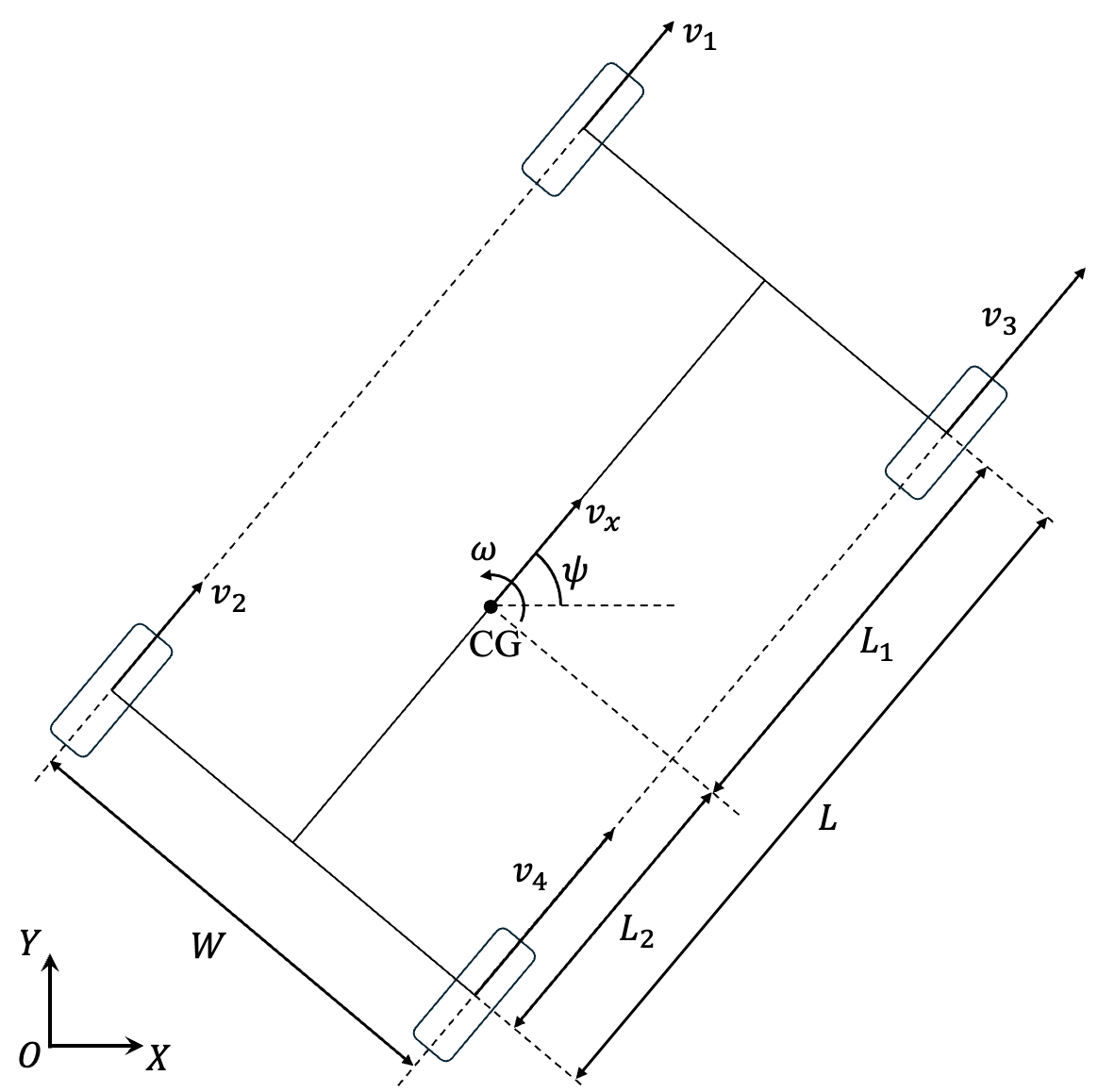}
    \caption{Schematic of the follower vehicle based on the differential-speed steering kinematic model.}
    \label{fig:vehicle-4wd}
\end{figure}

The full chassis control of the 4WD follower vehicle is achieved by independently commanding the desired wheel speeds of all four wheels. Steering is realized through the speed differential between the left and right wheels, producing a turning response equivalent to that generated by an effective steering angle of $\delta_\mathrm{obj}$.
As shown in Figure~\ref{fig:vehicle-4wd}, given the objective longitudinal velocity $v_{x,\mathrm{obj}}$ and steering angle $\delta_\mathrm{obj}$, the four wheel speed commands, $v_{1,\mathrm{obj}}$, $v_{2,\mathrm{obj}}$, $v_{3,\mathrm{obj}}$, and $v_{4,\mathrm{obj}}$, can be calculated using the Inverse Differential Speed Steering kinematics model~\cite{diffsteering}:
\begin{equation}
    v_{1,\mathrm{obj}}=v_{2,\mathrm{obj}} = v_{x,\mathrm{obj}} \left(1 - \frac{W}{2L}\tan\delta_\mathrm{obj}\right)
\end{equation}
\begin{equation}
    v_{3,\mathrm{obj}} = v_{4,\mathrm{obj}} = v_{x,\mathrm{obj}} \left(1 + \frac{W}{2L}\tan\delta_\mathrm{obj}\right)
\end{equation}

The vehicle’s estimated longitudinal velocity $\hat v_x$ and yaw rate $\hat \omega$ are obtained from wheel-speed measurements acquired through the motor encoders on all four wheels, based on the differential-speed steering kinematic model~\cite{diffsteering}:
\begin{equation}
    \hat v_x = \frac{\hat v_1 + \hat v_2 + \hat v_3 + \hat v_4}{4}
    \label{eq:vx_follower}
\end{equation}
\begin{equation}
    \hat \omega = \frac{-\hat v_1 - \hat v_2 + \hat v_3 + \hat v_4}{2W}
\end{equation}
where $\hat v_1$, $\hat v_2$, $\hat v_3$, and $\hat v_4$ are the measured wheel speeds and $W$ is the track-width.

Based on the hardware architecture and kinematic characteristics described above, motion control of these scaled vehicles is formulated in terms of two primary command variables: the desired longitudinal velocity and the desired steering input. These high-level control objectives must be generated in real time according to the vehicle states, tracking errors, and platoon coordination requirements. Accordingly, the following section presents the controller design used to compute these command variables for platooning operations.

\section{Platooning Leader-Follower Control Strategy}
To achieve robust vehicle platooning performance, the control system is decomposed into longitudinal and lateral control modules. The longitudinal controller regulates vehicle speed to maintain accurate and stable inter-vehicle spacing relative to the preceding vehicle. The lateral controller determines the steering command to ensure accurate path tracking and lateral alignment during maneuvers such as lane changes and turning.

\subsection{Longitudinal Control}
The longitudinal controller is implemented using a Proportional–Integral–Derivative (PID) control strategy, which seeks to minimize the error between the desired and measured inter-vehicle distance. 
The selection of a PID controller for longitudinal control is motivated by its simplicity, low computational cost, and well-established effectiveness in real-time automotive applications. It requires minimal system modeling and is well-suited for embedded implementations with strict timing constraints. Furthermore, since the primary objective of this study is the comparative evaluation of lateral control algorithms, employing a conventional and stable PID-based longitudinal controller ensures consistent velocity regulation without introducing additional variability. This design choice isolates the longitudinal dynamics and allows the analysis to focus on the performance differences among the lateral control strategies.

The PID control law is composed of three superimposed components: the proportional term provides an immediate response to the current error, the integral term accumulates past errors to eliminate steady-state offset, and the derivative term accounts for the rate of change of the error to anticipate future behavior~\cite{pid,pp-huang}. Its control law is given by:
\begin{equation}
    e_x(t) = d_{\text{measure}}(t) - d_{\text{goal}}(t)
\end{equation}
\begin{equation}
    v_{\text{des}}(t) = K_p \, e_x(t) + K_i \int_{0}^{t} e_x(t) \, dt + K_d \frac{d}{dt}e_x(t)
\end{equation}
where $e_x$ is the inter-vehicle distance tracking error, $d_{\text{measure}}$ is the measured inter-vehicle distance, $d_{\text{goal}}$ is the desired inter-vehicle following distance, $K_p$ is the proportional gain, $K_i$ is the integral gain, and $K_d$ is the derivative gain. 

\subsection{Lateral Control}
Geometric path tracking controllers are among the most widely used approaches for lateral vehicle control in robotics and autonomous driving applications. These methods exploit the geometric relationship between two vehicles to produce compact and intuitive steering control laws \cite{Snider2009}. Rather than relying on full vehicle dynamic models, geometric controllers operate on positional and heading error quantities measured directly from the sensors, making them well-suited for real-time implementation \cite{Jung2025}. Among the various geometric methods that have been proposed in the literature, Pure Pursuit and the Stanley controller remain the two most prominent and widely studied approaches. In this study, both of them are developed and compared for platoon leader-follower applications.

\subsubsection{Pure Pursuit Controller}
\begin{figure}[ht]
    \centering
    \includegraphics[width=0.65\linewidth]{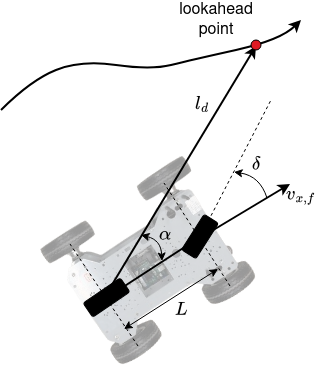}
    \caption{Schematic of the Pure Pursuit control method.}
    \label{fig:pp-schematics}
\end{figure}

In a vehicle-following context, rather than tracking a predefined static path, the Pure Pursuit controller directly uses the real-time position of the lead vehicle as the target point. The steering angle is determined from the curvature of the circular arc that guides the follower’s rear axle to the current position of the lead vehicle. As shown in Figure~\ref{fig:pp-schematics}, the distance between the follower’s rear axle and the lead vehicle is defined as the look-ahead distance $l_d$~\cite{Snider2009}. 
Based on the law of sines and the kinematic bicycle model, the steering command is given by \cite{park-pp,pp-YANG}:
\begin{equation}
    \delta(t) = \tan^{-1}\left(\frac{2L\sin(\alpha)}{l_d(t)}\right)
\end{equation}
\begin{equation}
    l_d(t) = K_{l_d} \cdot v_{x,f}(t)
\end{equation}
where $L$ is the vehicle wheelbase, $l_d$ is the lookahead distance, $K_{l_d}$ is the lookahead gain, and $\alpha$ is the Line-of-Sight (LOS) angle between the follower's heading direction and the direction pointing from the follower toward the lead vehicle. 
The lookahead distance $l_d$ is typically scaled as a linear function of the desired follower longitudinal velocity $v_{x,f}$. A large lookahead distance produces smooth but imprecise tracking and may cause corner-cutting, while a small lookahead improves accuracy at the cost of steering oscillations and potential instability \cite{pp-huang,pp-YANG,Snider2009}.

\subsubsection{Stanley Controller}
\begin{figure}[ht]
    \centering
    \includegraphics[width=0.85\linewidth]{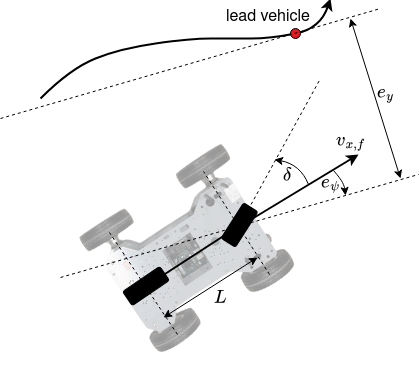}
    \caption{Schematic of the Stanley control method.}
    \label{fig:st-schematics}
\end{figure}

The Stanley controller provides an alternative geometric lateral control strategy for vehicle-following applications. As shown in Figure~\ref{fig:st-schematics}, it uses the center of the follower’s front axle as the reference point and simultaneously compensates for both heading misalignment and lateral offset relative to the real-time position of the lead vehicle~\cite{Snider2009,stanley-li}:
\begin{equation}
    \delta(t) = e_\psi(t) + \tan^{-1}\left(\frac{K_y \cdot e_y(t)}{v_{x,f}(t)+\varepsilon_v}\right)
\end{equation}
where $e_\psi$ is the heading error between the follower and lead vehicle, $v_{x,f}$ is the longitudinal velocity of the follower, $e_y$ is the lateral distance between the front axles of the follower and lead vehicle (also referred as the crosstrack error), $K_y$ is the crosstrack error gain, $\varepsilon_v$ is the regularization constant for mitigating abrupt steering variations and ensure smoother control action. 
The heading angle term steers the follower to align with the lead vehicle's heading, while the crosstrack term corrects lateral offset in a manner inversely proportional to speed, improving stability at higher following velocities \cite{sunstanley}. For small lateral offsets, the controller exhibits exponential convergence characteristics \cite{stanley-li}. The regularization constant $\varepsilon_v$ prevents numerical instability at very low speeds.

\section{Experimental Studies}
To evaluate the lateral control performance of the two geometric path-tracking controllers in a vehicle platoon scenario, the algorithms were implemented on the scaled multi-vehicle platform introduced in Section~\ref{platform}. Experiments were conducted on a flat indoor roadway with clearly defined lane boundaries.

\subsection{Vehicle Configuration}
\begin{figure}[ht]
    \centering
    \includegraphics[width=\linewidth]{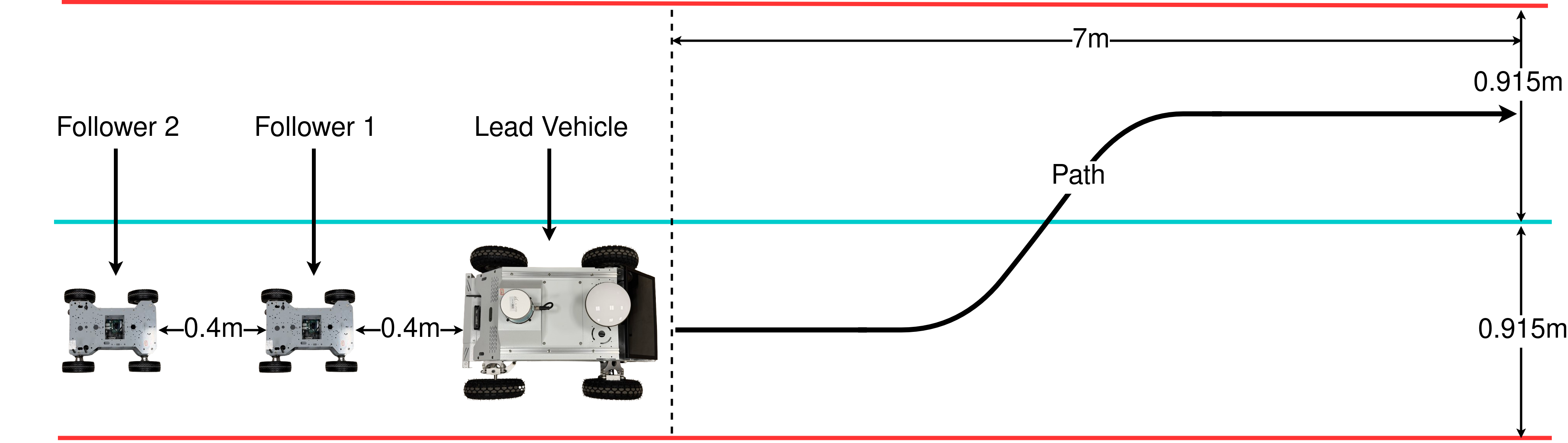}
    \caption{Experiment setup.}
    \label{fig:ExperimentSetup}
\end{figure}

The experimental platform consisted of three vehicles arranged in a convoy: one lead vehicle and two follower vehicles. The lead vehicle was pre-programmed to execute a single lane change maneuver, transitioning from the right lane to the left lane at a fixed speed of 0.2 m/s, as illustrated in Fig.~\ref{fig:ExperimentSetup}. 

\begin{figure}[ht]
    \centering
    \includegraphics[width=\linewidth]{IMRL_3_cars_2.png}
    \caption{Scaled multi-vehicle platform with ArUco markers demonstrating leader-follower platooning in an indoor corridor environment.}
    \label{fig:ArUco_on_car}
\end{figure}

To enable real-time estimation of the relative position between the lead and follower vehicles, ArUco markers were used for vision-based perception in this study because of their low computational cost, ease of deployment, and robust pose estimation capability under varying lighting conditions. ArUco markers are square fiducial markers designed for robust and efficient visual detection and pose estimation in computer vision systems, and are widely implemented in OpenCV. Each marker consists of a high-contrast black border enclosing a binary grid that encodes a unique identifier, allowing for reliable detection and discrimination among multiple markers~\cite{opencv_aruco_detection}.

Each follower vehicle uses its front-facing camera to detect the ArUco marker (ID: 582, size: 10 cm × 10 cm) mounted on the rear of the preceding vehicle, as shown in Fig.~\ref{fig:ArUco_on_car}. Detection of the ArUco marker provides real-time estimation of the relative distance and heading between consecutive vehicles, which are used as feedback signals for the longitudinal and lateral controllers. 
Both follower vehicles respond to the lane-change maneuver of the preceding vehicle using only onboard ArUco marker-based perception, without vehicle-to-vehicle (V2V) communication. The experiment is repeated for each lateral control method under identical environmental and initial conditions to ensure a fair comparison.

\subsection{Controller Configuration}
All follower vehicles utilize identical longitudinal PID controllers for inter-vehicle distance regulation. The lateral control strategy is varied between experimental rounds: Pure Pursuit is applied in the first round, and the Stanley method in the second. This design facilitates a controlled comparison of the performance between PID-Pure Pursuit and PID-Stanley control architectures.

The PID controller parameters for the follower vehicles are selected as $K_p$ = 1.5, $K_i$ = 0.3, and $K_d$ = 0. The desired inter-vehicle spacing is set to 0.5 m. For the Pure Pursuit controller, the look-ahead distance is chosen as 0.3 m to ensure stable steering behavior at low speeds~\cite{park-pp}. For the Stanley controller, the crosstrack gain is set to $k_y$ = 0.4, and the regularization constant $\varepsilon_v$ is assigned values of 0.001 m/s for left turns and -0.001 m/s for right turns.

\subsection{Round 1: PID and Pure Pursuit}
\begin{figure}[ht]
    \centering
    \includegraphics[width=\linewidth]{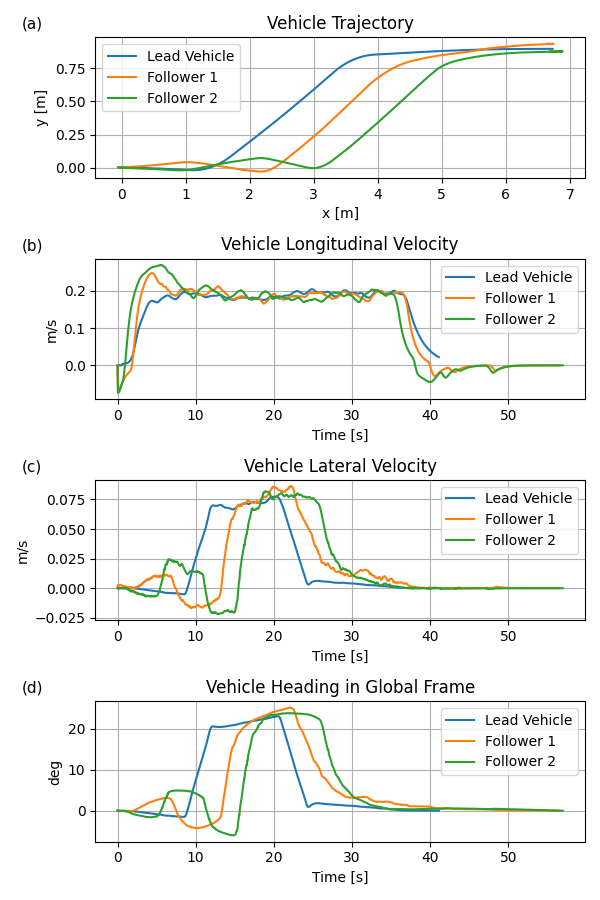}
    \caption{Experimental results using PID (longitudinal) and Pure Pursuit (lateral) controllers.}
    \label{fig:pp-results}
\end{figure}

The Pure Pursuit controller successfully guides all three vehicles through the lane change maneuver, with all vehicles converging to a steady-state lateral position of approximately 0.88-0.92 m as shown in Figure~\ref{fig:pp-results}(a). Prior to the lane change, a small lateral deviation is observed during the initial straight-line phase ($x$ = 0-2 m), caused by an initial heading misalignment that induces a brief corrective response before the vehicles re-align with the reference path. Follower~2 exhibits the largest transient undershoot of approximately 0.04~m near $x$ = 2.5 m.

Longitudinally, all vehicles converge to a cruising speed of approximately 0.2 m/s by $t$ = 10 s, as shown in Figure~\ref{fig:pp-results}(b), with oscillations of $\pm$0.02-0.03 m/s persisting during the cruise phase. An overshoot is observed at $t$ = 4-9 s due to the delayed response of the followers relative to the lead vehicle. The follower temporarily exceeds the steady-state velocity to compensate for the initial spacing error and achieve convergence. A transient undershoot of approximately -0.05~m/s is observed in Follower 2 near $t$ = 0-2 s, consistent with the initial heading correction.

The heading correction transients are relatively well attenuated in the velocity and yaw profiles. As shown in Figures~\ref{fig:pp-results}(c) and (d), the negative lateral velocity excursion in Follower~2 remains below -0.018 m/s near $t$ = 13 s, and the corresponding yaw excursion is limited to approximately -6° near $t$ = 14 s, both recovering gradually before the lane change commences. The lead vehicle produces a lateral velocity peak of approximately 0.07 m/s and a peak yaw angle of approximately 20-21° near $t$ = 20 s. Followers exhibit delayed and amplified responses, with lateral velocity peaks reaching 0.085 m/s and yaw peaks of up to 25°, indicative of string amplification along the platoon.

\subsection{Round 2: PID and Stanley}
\begin{figure}[ht]
    \centering
    \includegraphics[width=\linewidth]{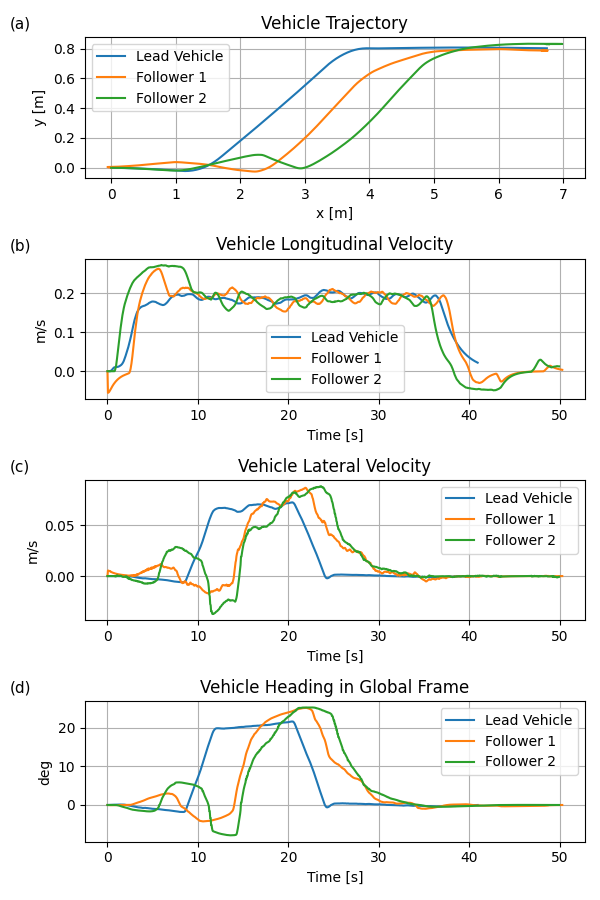}
    \caption{Experimental results using PID (longitudinal) and Stanley (lateral) controllers.}
    \label{fig:stanley-results}
\end{figure}

The Stanley controller similarly achieves stable lane change tracking, with all vehicles transitioning from $y$ = 0 m to a steady-state lateral position, as shown in Figure~\ref{fig:stanley-results}(a). The initial heading misalignment produces a comparable corrective deviation during the straight-line phase, however, the resulting transients are visibly more pronounced relative to the Pure Pursuit controller. Follower 2's lateral position undershoot remains consistent at approximately -0.04 m near $x$ = 2.5 m, suggesting that the position-level correction is comparable between the two lateral controllers.

Longitudinal velocity profiles are nearly identical to those of the Pure Pursuit controller, with all vehicles reaching approximately 0.2 m/s and the Follower 2 exhibiting a similar initial undershoot near $t$ = 0-2 s, as shown in Figure~\ref{fig:stanley-results}(b).

The key distinction appears in the velocity and yaw transients. The negative lateral velocity excursion in Follower 2 reaches approximately -0.038 m/s near $t$ = 13 s, as shown in Figure~\ref{fig:stanley-results}(c), accompanied by a negative yaw excursion of approximately -7° near $t$ = 14 s, as shown in Figure~\ref{fig:stanley-results}(d), before both recover abruptly as the lane change commences. These values are notably larger than those observed under Pure Pursuit, reflecting the Stanley controller's sensitivity to instantaneous heading error. The lead vehicle's lateral velocity and yaw angle peaks are consistent with the Pure Pursuit controller at approximately 0.07 m/s and 20-21°, respectively. Followers exhibit delayed and amplified peaks of up to 0.090~m/s in lateral velocity and 25° in yaw, confirming that string amplification is present regardless of controller type.

\subsection{Discussion}
Both the Stanley and Pure Pursuit lateral controllers successfully enabled the platoon to complete the lane-change maneuver while maintaining stable tracking throughout the test. However, clear differences in transient behavior were consistently observed.

The most notable distinction occurred during the pre-maneuver phase, when disturbances from the lead vehicle propagated through the platoon. The Stanley controller responded more aggressively because it relies on instantaneous tracking errors, resulting in amplified oscillations, particularly for Follower 2. In contrast, the Pure Pursuit controller generated smoother steering actions through its look-ahead mechanism, leading to improved disturbance attenuation.

String amplification was observed under both controllers, as reflected by increasing lateral velocity and yaw peaks along the platoon. Although the amplification remained bounded in this study, it highlights an inherent limitation of the predecessor-following structure and could become more pronounced at higher speeds or with shorter inter-vehicle spacing.

The longitudinal velocity response with a PID controller showed minimal sensitivity to the choice of lateral controller, indicating that the observed performance differences were primarily driven by the lateral control strategies.

Overall, the Pure Pursuit controller provided a better balance between tracking accuracy and transient smoothness in this experiment, making it more suitable for platoon lane-change scenarios under the conditions considered.

\section{Future Work}
After investigating PID, Pure Pursuit, and Stanley control strategies, future work will expand to model-based optimal control approaches such as Model Predictive Control (MPC). Building upon the current framework, where longitudinal and lateral controllers are designed independently, future advanced methods will jointly consider vehicle dynamics in both the longitudinal and lateral directions within an integrated control architecture. In addition to tracking the trajectory of the preceding vehicle, future controllers will explicitly incorporate relative speed and inter-vehicle state information to improve platoon safety, disturbance rejection, and string stability. Further studies will also examine controller performance under communication delays~\cite{ge2023communication}, sensor uncertainty, and more complex maneuvers such as merging, obstacle avoidance, and cooperative lane changes~\cite{yang2026collaborative}.

\section*{Acknowledgment}
The authors gratefully acknowledge financial support from ResearchNB under Grant ASF2026-020, as well as technical support provided by WHEELTEC (Dongguan) Co., Ltd.

\bibliographystyle{IEEEtran}
\bibliography{bibfile}

\end{document}